\documentclass{article}

\usepackage{amsmath,amssymb,amsthm}
\usepackage{tikz}
\usepackage{color}
\usepackage[toc]{appendix}
\usepackage{graphicx}
\usepackage{fancyhdr}
\usepackage{enumitem}
\usepackage{bbm}
\usepackage{parskip}
\usepackage{float}
\usepackage{chngpage}
\usepackage{calc}
\usepackage{bigints}
\usepackage{array}
\usepackage{booktabs}
\usepackage{rotating}
\usepackage{multirow}
\usepackage{adjustbox}
\usepackage{tabularx}
\usepackage{verbatim}
\usepackage{mathtools}
\usepackage{ragged2e}
\usepackage[makeroom]{cancel}
\usepackage{caption}
\usepackage{hyperref}
\usepackage{caption}
\usepackage{subcaption}
\usepackage{appendix}
\usepackage{pgfplots}
\pgfplotsset{compat=1.16}
\usepackage[english]{babel}
\usepackage{hyphenat}
\usepackage[makeindex]{imakeidx}
\usetikzlibrary{datavisualization}
\usetikzlibrary{matrix}
\usetikzlibrary{datavisualization.formats.functions}

\newtheorem{theorem}{Theorem}[section]

\newtheorem{remark}[theorem]{Remark}
\newtheorem{assumption}[theorem]{Assumption}

\makeatletter
\def\section{\@startsection {section}{1}{\z@}{3.25ex plus 1ex minus
		.2ex}{1.5ex plus .2ex}{\large\bf}}
\def\subsection{\@startsection{subsection}{2}{\z@}{3.25ex plus 1ex minus
		.2ex}{1.5ex plus .2ex}{\normalsize\bf}}
\@addtoreset{equation}{section} 
\makeatother

\title{A note on diffusion limits for stochastic gradient descent}

\author{Alberto Lanconelli\thanks{Dipartimento di Scienze Statistiche Paolo Fortunati, Università di Bologna, Bologna, Italy. \textbf{e-mail}: alberto.lanconelli2@unibo.it} \and Christopher S. A. Lauria\thanks{Dipartimento di Scienze Statistiche Paolo Fortunati, Università di Bologna, Bologna, Italy. \textbf{e-mail}: christopher.lauria2@unibo.it}}

\date{\today}

\begin{document}
	
\maketitle
	
\bigskip
	
\begin{abstract}
In the machine learning literature stochastic gradient descent has recently been widely discussed for its purported implicit regularization properties. Much of the theory, that attempts to clarify the role of noise in stochastic gradient algorithms, has approximated stochastic gradient descent by a stochastic differential equation with Gaussian noise. We provide a rigorous theoretical justification for this practice that showcases how the Gaussianity of the noise arises naturally.
\end{abstract}
	
	
	
\allowdisplaybreaks
	
\section{Introduction}\label{intro}
\emph{Stochastic gradient descent} (SGD) (\cite{Robbins}, \cite{Nemiro}) is an optimization algorithm that has found a great deal of success in applied settings, especially in the machine learning framework \cite{tsypkin1971}, \cite{Bottou98}, \cite{Bottou2018}.
Its success has spurred the creation of many variants, \cite{rumelhart1986learning}, \cite{kingma2014adam}, \cite{duchi2011adaptive}, and its properties have been studied in a number of ways, \cite{Netrapalli}, \cite{Gower}, \cite{Nguyen2018}, \cite{moulines2011non}, \cite{toulis2017asymptotic}. \\
A widespread setting where SGD is used concerns minimization problems of the form
\begin{align}\label{1*}
\min_{x \in \mathbb{R}^d} f(x) := \min_{x \in \mathbb{R}^d} \frac{1}{n}\sum_{i=1}^n f_i(x)
\end{align}
where $f_i: \mathbb{R}^d \rightarrow \mathbb{R}$ for $i = 1, \dots, n$ represent some given \emph{loss functions}.
The SGD recursive iteration, in its simplest form, can be written as
\begin{align} \label{sgd}
x^{(\eta)}_{0}=x_0\in\mathbb{R}^d\quad\mbox{ and }\quad x^{(\eta)}_{k+1} = x^{(\eta)}_{k} - \eta \nabla f_{\gamma_{k+1}}(x^{(\eta)}_{k}),\quad k\geq 0
\end{align}
where $\eta$ is a positive constant, usually named \emph{learning rate}, while $\{ \gamma_k\}_{k \in \mathbb{N}}$ is a sequence of independent discrete uniform random variables on the set $\{1, 2, \dots, n\}$. Here we stress through the superscript $\eta$ in $x^{(\eta)}_{k}$ the crucial dependence of the iterative scheme on the size of the learning rate. 

Recently, the iteration scheme \eqref{sgd} has been approximated, in the limit as $\eta$ tends to zero, by certain continuous time analogues which can be formalized in terms of stochastic differential equations (SDE), see for instance \cite{Mandt2015} and \cite{chaudhari2018}. This permits an asymptotic analysis of stochastic gradient algorithms. Moreover, the continuous time treatment also allows  the application of optimal control theory to study the problems of adaptive hyper-parameter adjustments \cite{Li2017}. \\
Although the SDE approximation to SGD has been used throughout the literature, there has, to the authors knowledge, only been one partially rigorous formalization of it, given by \cite{Li2017} and \cite{Li2019}, where stochastic gradient algorithms are shown to be approximated in distribution by SDEs driven by, appropriately chosen, Gaussian noise. However, some authors have questioned the traditional assumption that SGD noise is Gaussian \cite{Nguyen2019}, \cite{Simsek}, see also \cite{li2021} . 

The aim of this note is to prove that the Gaussianity of the noise driving the SDE proposed in \cite{Li2017} and \cite{Li2019} can be rigorously derived, thus providing an additional theoretical justification for its pervasive use in the literature. Our approach, inspired by a similar setting in the econometric literature \cite{Nelson} (see also \cite{Corradi} and \cite{Flandoli}), utilizes a general theorem from \cite{stroock1997} which provides sufficient conditions for a sequence of discrete time Markov processes to convergence in distribution to an It\^o SDEs driven by Brownian motion.

\section{Heuristic derivation of the diffusion limit and statement of the main result}

Before stating our main result we summarize the procedure proposed by the authors in \cite{Li2017} and \cite{Li2019} to derive an SDE from the iterative scheme \eqref{sgd}. The main idea is to treat the learning rate $\eta$ as the time step of a discretization scheme for a continuous time stochastic process. To this aim the iterative rule \eqref{sgd} will be more conveniently written as 
\begin{align}\label{sgd*}
x^{(\eta)}_{\eta(k+1)} = x^{(\eta)}_{\eta k} - \eta \nabla f_{\gamma_{k+1}}(x^{(\eta)}_{\eta k}),
\end{align}
so that the difference between the lower indexes of $x^{(\eta)}_{\eta(k+1)}$ and $x^{(\eta)}_{\eta k}$ agrees with the proportionality constant in front of the gradient $\nabla f_{\gamma_{k+1}}(x^{(\eta)}_{\eta k})$. \\
The two preparatory steps used in \cite{Li2019} for relating \eqref{sgd*} to an SDE are: 
\begin{itemize}
\item Rewriting \eqref{sgd*} (recalling the definition of $f$ in \eqref{1*}) as
\begin{align} \label{sgd2}
x^{(\eta)}_{\eta(k+1)} &= x^{(\eta)}_{\eta k} - \eta \nabla f_{\gamma_{k+1}}(x^{(\eta)}_{\eta k})\nonumber\\
&=x^{(\eta)}_{\eta k} - \eta \left(\nabla f_{\gamma_{k+1}}(x^{(\eta)}_{\eta k})-\nabla f(x_{\eta k}^{(\eta)})+\nabla f(x_{\eta k}^{(\eta)})\right)\nonumber\\
&=x^{(\eta)}_{\eta k} -\nabla f(x_{\eta k}^{(\eta)})\eta+\left(\nabla f(x_{\eta k}^{(\eta)})-\nabla f_{\gamma_{k+1}}(x^{(\eta)}_{\eta k})\right)\eta\nonumber\\
&=x^{(\eta)}_{\eta k} -\nabla f(x_{\eta k}^{(\eta)})\eta+V_{\eta k}^{(\eta)}\sqrt{\eta},
\end{align}
where we set
\begin{align*}
V_{\eta k}^{(\eta)}:=(\nabla f(x_{\eta k}^{(\eta)})-\nabla f_{\gamma_{k+1}}(x^{(\eta)}_{\eta k}))\sqrt{\eta},\quad k\geq 0.
\end{align*}
Notice that $V_{\eta k}^{(\eta)}$ is a $d$-dimensional random vector with conditional mean
\begin{align*}
\mathbb{E}[V_{\eta k}^{(\eta)}|\mathcal{F}^{(\eta)}_{\eta k}]=0
\end{align*}
and conditional covariance matrix
\begin{align*}
\mathbb{E}\left[V_{\eta k}^{(\eta)}(V_{\eta k}^{(\eta)})^T|\mathcal{F}^{(\eta)}_{\eta k}\right]&=\frac{\eta}{n}\sum_{i=1}^n(\nabla f(x_{\eta k}^{(\eta)})-\nabla f_{i}(x^{(\eta)}_{\eta k}))(\nabla f(x_{\eta k}^{(\eta)})-\nabla f_{i}(x^{(\eta)}_{\eta k}))^T\\
&=\eta\Sigma(x_{\eta k}^{(\eta)}),
\end{align*}
where we denoted
\begin{align}\label{sigma}
\Sigma(x):=\frac{1}{n}\sum_{i=1}^n(\nabla f(x)-\nabla f_{i}(x)(\nabla f(x)-\nabla f_{i}(x)^T,
\end{align}
while $\mathcal{F}_{\eta k}^{(\eta)}$ stands for the $\sigma$-algebra generated by the random vectors $x^{(\eta)}_0,x_\eta^{(\eta)},...,x_{\eta k}^{(\eta)}$ for $k\geq 0$. 
\item Replacing $V_{\eta k}^{(\eta)}$ with
\begin{align*}
\overline{V_{\eta k}^{(\eta)}}:=(\nabla f(x_{\eta k}^{(\eta)})-\nabla f_{\gamma_{k+1}}(x^{(\eta)}_{\eta k}))\sqrt{\overline{\eta}},\quad k\geq 0,
\end{align*}
where $\overline{\eta}$ plays the role of a new parameter independent of $\eta$; observe that the conditional covariance matrix of $\overline{V^{(\eta)}_{\eta k}}$ is now
\begin{align}\label{frozen covariance}
	\mathbb{E}\left[\overline{V_{\eta k}^{(\eta)}}(\overline{V_{\eta k}^{(\eta)}})^T|\mathcal{F}^{(\eta)}_{ \eta k}\right]=\overline{\eta}\Sigma(x_{\eta k}^{(\eta)}).
\end{align}
With such a substitution equation \eqref{sgd2} reads
\begin{align} \label{sgdf}
x^{(\eta)}_{\eta(k+1)} &=  x^{(\eta)}_{\eta k}-\nabla f(x^{(\eta)}_{\eta k})\eta +  \overline{V^{(\eta)}_{\eta k}}\sqrt{\eta},\quad k\geq 0.  
\end{align}
\end{itemize}
At this point, the authors in \cite{Li2019} take a further step and replace the term $\overline{V^{(\eta)}_{\eta k}}\sqrt{\eta}$ in \eqref{sgdf} with a Gaussian random vector having the same mean and covariance as $\overline{V^{(\eta)}_{\eta k}}\sqrt{\eta}$; more precisely, they consider the random vector
\begin{align*}
	\left(\overline{\eta}\Sigma(x_{\eta k}^{(\eta)})\right)^{\frac{1}{2}}(B_{(k+1)\eta}-B_{k\eta}),
\end{align*}  
with $\{B_t\}_{t\geq 0}$ being a $d$-dimensional standard Brownian motion, which transforms \eqref{sgdf} into
\begin{align*}
x^{(\eta)}_{\eta(k+1)} &=  x^{(\eta)}_{\eta k}-\nabla f(x^{(\eta)}_{\eta k})\eta +  \left(\overline{\eta}\Sigma(x_{\eta k}^{(\eta)})\right)^{\frac{1}{2}}(B_{(k+1)\eta}-B_{k\eta}),\quad k\geq 0. 
\end{align*}
This equation corresponds to the Euler scheme (see for instance \cite{KP}) with step size $\eta$ for the $d$-dimensional It\^o's SDE
\begin{align}\label{SDE}
dX_t=-\nabla f(X_t)dt+\left(\overline{\eta}\Sigma(X_t)\right)^{\frac{1}{2}}dB_t,\quad t\geq 0.
\end{align}
The SDE in \ref{SDE} is then taken as a candidate for the continuous time limit of the iteration scheme \eqref{sgd*}. Subsequently \cite{Li2019} prove that for any $\mathtt{T}>0$, $k\in\{0,...,[\mathtt{T}/\bar{\eta}]\}$ and any polynomially bounded function $g:\mathbb{R}^d\to\mathbb{R}$ one has
\begin{align*}
|\mathbb{E}[g(X_{\bar{\eta}k})]-\mathbb{E}[g(x_k^{(\bar{\eta})})]|\leq C\bar{\eta}.
\end{align*}
The solution to \eqref{SDE} is thus interpreted as a weak asymptotic approximation, as $\bar{\eta}$ tends to zero, of the SGD update $x_k^{(\bar{\eta})}$ in \eqref{sgd}.\\
The aim of this note is to show that \eqref{SDE} can be rigorously derived from \eqref{sgdf}, entailing the intrinsic Gaussian nature of the noise. To do so we will invoke a general functional limit theorem for discrete time Markov processes established in \cite{stroock1997}. Our result reads as follows.
\newpage
\begin{theorem} \label{mainthm}
	Assume that:
	\begin{enumerate}
		\item the gradients $\nabla f_i$, $i=1,...,n$ are Lipschitz continuous;
		\item the covariance matrix $\Sigma$ defined in \eqref{sigma} obeys the bound  
	\begin{align}\label{condition on covariance}
	\max_{j\in\{1,...,d\}}\left|\langle\theta,\partial^2_{x_j}\Sigma(x)\theta\rangle\right|\leq\lambda_0\|\theta\|^2,\quad \theta\in\mathbb{R}^d 
	\end{align}
	for all $x\in\mathbb{R}^d$ (here, the differential operator $\partial^2_{x_j}$ acts on the matrix $\Sigma(x)$ entrywise). 
	\end{enumerate}
	Then, the discrete time stochastic process $\{x^{(\eta)}_{\eta k}\}_{k\geq 0}$ defined recursively in \eqref{sgdf} converges in distribution, as $\eta$ tends to zero, to the unique weak solution of the It\^o SDE \eqref{SDE}.
\end{theorem}

\begin{remark}
The assumption concerning the Lipschitz continuity of the gradients $\nabla f_i$, $i=1,...,n$ is natural considering that the convergence of SGD is canonically proven under such an assumption. On the other hand, the requirement on the covariance matrix $\Sigma$ is needed to ensure a well behaved limiting equation \eqref{SDE}. 
\end{remark}

\section{Proof of Theorem \ref{mainthm}}

The proof of our main result consists in an application of a general functional limit theorem stated in \cite{Stroock}. Here, we utilize a slightly different version, proposed in \cite{Nelson}, which we further simplify to match our framework. The next section is devoted to its rigorous statement.

\subsection{A general functional limit theorem}

Let $\left\{x^{(\eta)}_{\eta k}\right\}_{k\geq 0}$ be a $d$-dimensional discrete time Markov process defined on a complete probability space $(\Omega,\mathcal{F},\mathbb{P})$ and  indexed by the positive real number $\eta$. We assume that 
\begin{align*}
\mathbb{P}\left(x^{(\eta)}_0=x_0\right)=1,\quad\mbox{ for all $\eta>0$}
\end{align*}
where $x_0$ is a deterministic $d$-dimensional vector. For $\eta>0$ and $k\geq 0$ we write $\mathcal{F}^{(\eta)}_{k \eta}$ to represent the $\sigma$-algebra generated by the random vectors $x_0, x_{\eta}^{(\eta)},x_{2\eta}^{(\eta)},...,x_{k\eta}^{(\eta)}$; the Markov property of $\left\{x^{(\eta)}_{\eta k}\right\}_{k\geq 0}$ yields
\begin{align*}
\mathbb{E}[\varphi(x^{(\eta)}_{\eta (k+1)})|\mathcal{F}^{(\eta)}_{k \eta}]=\mathbb{E}[\varphi(x^{(\eta)}_{\eta (k+1)})|x^{(\eta)}_{\eta k}]=\int_{\mathbb{R}^d}\varphi(y)\Pi^{(\eta)}_{\eta k}(x^{(\eta)}_{\eta k};dy),
\end{align*}
for any bounded and measurable $\varphi:\mathbb{R}^d\to\mathbb{R}$; here, $\Pi^{(\eta)}_{\eta k}$ denotes the transition kernel of the process $\left\{x^{(\eta)}_{\eta k}\right\}_{k\geq 0}$.
We now embed the discrete time process $\left\{x^{(\eta)}_{\eta k}\right\}_{k\geq 0}$ in a continuous time one, denoted $\{X_t^{(\eta)}\}_{t\geq 0}$, through the prescription
\begin{align}\label{continuous}
X_t^{(\eta)}:=x^{(\eta)}_{k\eta},\quad \mbox{ if }t\in [k\eta,(k+1)\eta[. 
\end{align}
The following set of assumptions will imply the weak convergence of $\{X_t^{(\eta)}\}_{t\geq 0}$ towards the solution of an It\^o SDE, as $\eta$ tends to zero.
\begin{assumption}\label{as1} 
There exist continuous functions $a:\mathbb{R}^d \times [0, \infty[\to S_{d}$, the space of $d \times d$ symmetric non negative definite matrices, and $b:\mathbb{R}^d \times [0, \infty[\to\mathbb{R}^d$ such that for all $R>0$ and $\mathtt{T}>0$ we get
\begin{align}\label{1}
&\lim_{\eta \rightarrow 0} \sup_{\| x \| \leq R,0\leq t\leq \mathtt{T}} \left\| \frac{\mathbb{E}[X^{(\eta)}_{t+\eta}-X^{(\eta)}_{t}\vert X^{(\eta)}_{t}=x]}{\eta}
  - b(x,t) \right\| = 0
  \end{align}
and
\begin{align}\label{2}
&\lim_{\eta \rightarrow 0} \sup_{\| x \| \leq R, 0\leq t\leq \mathtt{T}} \left\|\frac{\mathbb{E}[(X^{(\eta)}_{t+\eta}-X^{(\eta)}_{t})(X^{(\eta)}_{t+\eta}-X^{(\eta)}_{t})^T\vert X^{(\eta)}_{t}=x]}{\eta} - a(x,t) \right\|  = 0.
\end{align}
Moreover, there exists a positive $\delta$ such that for all $j=1,...,d$ we have
\begin{align}\label{3}
\lim_{\eta \rightarrow 0} \sup_{\| x \| \leq R, 0\leq t\leq \mathtt{T}} \frac{\mathbb{E}[|\langle X^{(\eta)}_{t+\eta}-X^{(\eta)}_{t},e_j\rangle|^{2+\delta}\vert X^{(\eta)}_{t}=x]}{\eta} = 0;
\end{align}
here, $\{e_1,...,e_d\}$ denotes the canonical basis of $\mathbb{R}^d$.
\end{assumption} 
 
\begin{assumption}\label{as2} 
	There exists a continuous mapping $\sigma:\mathbb{R}^d \times [0, \infty[\to S_{d}$ such that for all $x \in \mathbb{R}^d $ and $t\geq 0$ we have $ a(x,t) = \sigma(x,t) \sigma(x,t)^T$. 
\end{assumption}

\begin{assumption}\label{as4} 
	Weak uniqueness holds for the It\^o SDE
	\begin{align}\label{Ito}
	\begin{cases}
	dX_t=b(X_t,t)dt+\sigma(X_t,t)dB_t,& t>0;\\
	X_0=x_0.&
	\end{cases}
	\end{align}
Here, $\{B_t\}_{t\geq 0}$ denotes a standard $d$-dimensional Brownian motion.
\end{assumption}

We are now ready to state the following theorem taken from \cite{Nelson}. 

\begin{theorem} \label{thm1}
Under Assumptions \ref{as1}-\ref{as4} the continuous time stochastic process $\{X^{(\eta)}_t\}_{t\geq 0}$ converges in distribution, as $\eta$ tends to zero, to the unique weak solution $\{X_t\}_{t\geq 0}$ of the It\^o SDE \eqref{Ito}.
\end{theorem}

\subsection{Proof of the main result}

To apply Theorem \ref{thm1} we have to verify the validity of Assumptions \ref{as1}-\ref{as4} for $\{X_t^{(\eta)}\}_{t\geq 0}$ being defined by \eqref{continuous} with $\{x^{(\eta)}_{\eta k}\}_{k\geq 0}$ from \eqref{sgdf}.
\begin{itemize}
	\item \emph{Verification of Assumption \ref{as1}}: fix $t>0$ and assume that $k\eta\leq t<(k+1)\eta$ for some $k\in\mathbb{N}$. Then, according to \eqref{continuous} and \eqref{sgdf} we can write
\begin{align*}
X_{t+\eta}^{(\eta)}-X_t^{(\eta)}&=x^{(\eta)}_{(k+1)\eta}-x^{(\eta)}_{k\eta}\\
&=-\nabla f(x^{(\eta)}_{\eta k})\eta +  \overline{V^{(\eta)}_{\eta k}}\sqrt{\eta}
\end{align*}
and hence
\begin{align*}
\frac{\mathbb{E}[X^{(\eta)}_{t+\eta}-X^{(\eta)}_{t}\vert X^{(\eta)}_{t}=x]}{\eta}&=\frac{\mathbb{E}[-\nabla f(x^{(\eta)}_{\eta k})\eta +  \overline{V^{(\eta)}_{\eta k}}\sqrt{\eta}\vert x^{(\eta)}_{\eta k}=x]}{\eta}\\
&=\mathbb{E}[-\nabla f(x^{(\eta)}_{\eta k})\vert X^{(\eta)}_{t}=x]+\frac{\mathbb{E}[\overline{V^{(\eta)}_{\eta k}}\vert x^{(\eta)}_{\eta k}=x]}{\sqrt{\eta}}\\
&=\mathbb{E}[-\nabla f(x^{(\eta)}_{\eta k})\vert x^{(\eta)}_{\eta k}=x]\\
&=-\nabla f(x).
\end{align*}
Therefore, if we set $b(t,x):=-\nabla f(x)$ condition \eqref{1} is trivially satisfied. Moreover, a similar computation gives
\begin{align*}
&\frac{\mathbb{E}[(X^{(\eta)}_{t+\eta}-X^{(\eta)}_{t})(X^{(\eta)}_{t+\eta}-X^{(\eta)}_{t})^T\vert X^{(\eta)}_{t}=x]}{\eta}\\
&\quad=\frac{\mathbb{E}[(-\nabla f(x^{(\eta)}_{\eta k})\eta +  \overline{V^{(\eta)}_{\eta k}}\sqrt{\eta})(-\nabla f(x^{(\eta)}_{\eta k})\eta +  \overline{V^{(\eta)}_{\eta k}}\sqrt{\eta})^T\vert x^{(\eta)}_{\eta k}=x]}{\eta}\\
&\quad=\nabla f(x)(\nabla f(x))^T\eta+\mathbb{E}[\overline{V^{(\eta)}_{\eta k}}(\overline{V^{(\eta)}_{\eta k}})^T]\\
&\quad=\nabla f(x)(\nabla f(x))^T\eta+\overline{\eta}\Sigma(x).
\end{align*}
Hence, setting $a(x,t):=\overline{\eta}\Sigma(x)$ we satisfy assumption \eqref{2}. \\ Observe that the assumption of Lipschitz continuity of $\nabla f(x)$ entails the boundedness of $\nabla f(x)(\nabla f(x))^T$ on the compact sets $\|x\|\leq R$.\\
We now verify \eqref{3}: let $j\in\{1,...,d\}$ and $\delta>0$; then,
\begin{align*}
&\frac{\mathbb{E}[|\langle  X^{(\eta)}_{t+\eta}-X^{(\eta)}_{t},e_j\rangle|^{2+\delta}|X^{(\eta)}_{t}=x]}{\eta}\\
&\quad=\frac{\mathbb{E}[|-\partial_{x_j} f(x^{(\eta)}_{\eta k})\eta +(\partial_{x_j} f(x_{\eta k}^{(\eta)})-\partial_{x_j} f_{\gamma_{k+1}}(x^{(\eta)}_{\eta k}))\sqrt{\bar{\eta}}\sqrt{\eta}|^{2+\delta}|x^{(\eta)}_{\eta k}=x]}{\eta}\\
&\quad\leq2^{1+\delta}\frac{\mathbb{E}[|-\partial_{x_j} f(x^{(\eta)}_{\eta k})\eta|^{2+\delta} +|(\partial_{x_j} f(x_{\eta k}^{(\eta)})-\partial_{x_j} f_{\gamma_{k+1}}(x^{(\eta)}_{\eta k}))\sqrt{\bar{\eta}}\sqrt{\eta}|^{2+\delta}|x^{(\eta)}_{\eta k}=x]}{\eta}\\
&\quad\leq c_{\delta,\bar{\eta}}\left(|\partial_{x_j} f(x)|^{2+\delta}\eta^{1+\delta}+\mathbb{E}[|(\partial_{x_j} f(x_{\eta k}^{(\eta)})-\partial_{x_j} f_{\gamma_{k+1}}(x^{(\eta)}_{\eta k}))|^{2+\delta}|x^{(\eta)}_{\eta k}=x]\eta^{\frac{\delta}{2}}\right)\\
&\quad \leq c_{\delta,\bar{\eta}}\left(|\partial_{x_j} f(x)|^{2+\delta}\eta^{1+\delta}+2^{1+\delta}|\partial_{x_j} f(x)|^{2+\delta}\eta^{\frac{\delta}{2}}
+2^{1+\delta}\mathbb{E}[|\partial_{x_j} f_{\gamma_{k+1}}(x^{(\eta)}_{\eta k}))|^{2+\delta}|x^{(\eta)}_{\eta k}=x]\eta^{\frac{\delta}{2}}\right)\\
&\quad \leq c_{\delta,\bar{\eta}}\left(|\partial_{x_j} f(x)|^{2+\delta}\eta^{1+\delta}+2^{1+\delta}|\partial_{x_j} f(x)|^{2+\delta}\eta^{\frac{\delta}{2}}
+2^{1+\delta}\eta^{\frac{\delta}{2}}\frac{1}{n}\sum_{i=1}^n|\partial_{x_j} f_{i}(x)|^{2+\delta}\right).
\end{align*}
Here, in the first and third inequalities, we utilized the basic estimate $|a+b|^p\leq 2^{p-1}(|a|^p+|b|^p)$ while $c_{\delta,\bar{\eta}}$ stands for a positive constant whose value may vary from line to line.
Since the terms containing the partial derivatives of $f$ are bounded on the compact sets $\|x\|\leq R$ (given that the gradients $\nabla f_i$ are assumed to be Lipschitz continuous) we conclude that passing to the limit as $\eta$ tends to zero the last term above will converge to zero, thus making condition \eqref{3} hold true.
\item \emph{Verification of Assumption \ref{as2}}: we know from before that $a(x,t)=\overline{\eta}\Sigma(x)$; being a covariance matrix, $\Sigma(x)$ is symmetric and positive semi-definite thus the existence of a unique symmetric and positive semi-definite matrix $\sigma(x)$ such that $a(x,t)=\overline{\eta}\sigma(x)\sigma(x)$ is entailed. In addition, according to Theorem 5.2.3 in \cite{stroock1997} the bound in \eqref{condition on covariance} implies the global Lipschitz continuity of $\sigma(x)$.
\item \emph{Verification of Assumption \ref{as4}}: since $b(t,x)=-\nabla f(x)$ and $\sigma(x,t)=\left(\overline{\eta}\Sigma(x)\right)^{\frac{1}{2}}$, the assumption on the gradients $\nabla f_i$ together with \eqref{condition on covariance} yield the Lipschitz continuity for the coefficients of the It\^o SDE \eqref{SDE}, this entails strong (and hence weak) uniqueness.
\end{itemize}

\bibliographystyle{apalike}
\bibliography{bibfile}

\end{document}